\documentclass[cameraready]{Interspeech}
\usepackage{amsmath,amssymb}
\usepackage{graphicx}
\usepackage{booktabs}
\usepackage{multirow}
\usepackage{xcolor}
\usepackage{hyperref}
\usepackage{microtype}
\usepackage{enumitem}
\usepackage[T5]{fontenc}
\usepackage[utf8]{inputenc}
\usepackage[vietnamese,english]{babel}

\title{PiDA: Phonetically-Informed Data Augmentation \\for Robust Vietnamese Speech Translation}

\author[affiliation={1}]{Giang Son}{Nguyen}
\author[affiliation={1}]{Tung X.}{Nguyen}
\author[affiliation={1}]{Hieu Minh}{Truong}
\author[affiliation={1,2}]{Nhu}{Vo}
\author[affiliation={1,3}]{Wray}{Buntine}
\author[affiliation={1}]{Dung D.}{Le}

\address{
  $^1$ VinUniversity, Vietnam \\
  $^2$ University of Technology Sydney, Australia \quad  $^3$ Monash University, Australia
}
\email{\{son.ng, tung.nx, hieu.tm2, nhu.vd, wray.b, dung.ld\}@vinuni.edu.vn}

\newcommand{\vn}[1]{\textit{#1}}

\usepackage{comment}

\begin{document}
\maketitle

\begin{abstract}
Cascaded speech translation (ST) systems suffer from error propagation when Automatic Speech Recognition (ASR) outputs incorrect transcripts. We present the first systematic categorization of ASR errors for Vietnamese ST, classifying substitution errors by phonetic cause and quantifying their impact on downstream Neural Machine Translation (NMT) performance using Linear Mixed-Effects Modelling. We confirm that most ASR substitution errors arise from phonetic confusions rather than random noise, and that these phonetic errors significantly degrade ST quality. Motivated by this finding, we propose Phonetically-Informed Data Augmentation (PiDA), which generates ASR-like corruptions by substituting words with phonetically similar alternatives using phonetic word embeddings. Fine-tuning on a PiDA-augmented version of FLEURS Vietnamese–English improves translation of erroneous ASR outputs (up to +2.04 BLEU over standard fine-tuning) while also slightly improving clean-text performance.
\end{abstract}

\section{Introduction}

Speech translation (ST) converts spoken language directly into text in another language. Two paradigms dominate: end-to-end ST (E2E ST), which maps audio directly to translated text, and cascaded ST, which combines an Automatic Speech Recognition (ASR) system with a Neural Machine Translation (NMT) model. Despite progress in E2E ST, recent benchmarks show that cascaded systems often match or outperform them \cite{dabre-song-2024-nicts, yan-etal-2024-cmus, higuchi25_interspeech, le-duc-etal-2025-multimed-st}.

However, a key limitation of cascaded ST is \textit{error propagation}: ASR errors are passed to the NMT component, degrading translation quality. This effect is amplified by a training–inference mismatch, as NMT models trained on clean text must handle noisy ASR outputs at test time \cite{sperber2017toward,belinkov2018synthetic,cheng2018towards,di-gangi-etal-2019-robust,binici2025medsage}. On the Vietnamese–English subset of FLEURS \cite{conneau2023fleurs}, this mismatch results in 6.79 to 10.64 BLEU drops when translating ASR outputs instead of clean transcripts (Section~\ref{experimental_results}).

\begin{figure}[t]
\small
\setlength{\fboxsep}{3pt}%
\fbox{%
\begin{minipage}[t]{\dimexpr\columnwidth-2\fboxsep-2\fboxrule\relax}
\textbf{Qualitative Examples}

\vspace{0.3em}
\textbf{(1) Phonetically-Informed Data Augmentation (PiDA)}

\textit{Clean (Vi):} một \textbf{trong} những\dots \textbf{quy} tắc\dots \textbf{tây} ban nha  \\
\textit{PiDA (Vi):} một \underline{ngon} những\dots \underline{quê} tắc\dots \underline{thây} ban nha  

\vspace{0.2em}
{\scriptsize synthetic substitutions sampled from phonetically similar syllables}

\vspace{0.4em}
\textbf{(2) Robustness via FT with PiDA}

\textit{Reference (Vi):} vỡ ra với \textcolor{blue}{\textbf{ít}$_\text{little}$} hoặc không ảnh hưởng \\
\textit{ASR output (Vi):} vỡ ra với \textcolor{red}{\underline{ích}$_\text{useful}$} hoặc không ảnh hưởng

\vspace{0.2em}
\textit{Clean FT:} break up \textcolor{red}{usefully} or without effect (BLEU: 47.8) \\
\textit{PiDA FT:} break apart with \textcolor{blue}{little} or no effect (BLEU: 93.9)

\vspace{0.2em}
{\scriptsize FT with PiDA recovers intended meaning under phonetic confusion}
\end{minipage}%
}
\vspace{-0.6em}
\caption{Qualitative examples: (1) A sentence from the training set corrupted by PiDA; (2) A sentence from the test set where a model finetuned with PiDA data translates correctly despite ASR errors.}
\label{fig:pida-qual}
\end{figure}

A common approach to improving NMT robustness is to augment training data with noisy or corrupted input text \cite{sperber2017toward,belinkov2018synthetic}. However, \cite{cheng2018towards} show that robustness gains depend on the match between injected noise and real error distributions, raising a question: \textit{what kinds of errors should we inject?}

To answer this question, we conduct, to our knowledge, the first systematic analysis of ASR error patterns in Vietnamese speech translation. We categorize substitution errors by their phonetic causes, and quantify each category’s impact on translation quality using a linear mixed-effects model. Our analysis confirms that \textbf{the majority of ASR substitution errors arise from systematic phonetic confusions rather than random noise} and that \textbf{phonetic ASR errors meaningfully impact NMT performance}. 

This finding directly motivates our approach: since ASR errors are predominantly phonetic, synthetic augmentation should also be phonetic. We propose \textbf{P}honetically-\textbf{I}nformed \textbf{D}ata \textbf{A}ugmentation (\textbf{PiDA}) using phonetic embeddings \cite{zouhar-etal-2024-pwesuite, sperduti-nguyen-2025-pset}, which generates corruptions by substituting syllables with phonetically similar alternatives identified through embedding similarity. To our knowledge, this is the first work to use explicit phonetic embeddings to generate synthetic ASR-style training data for machine translation on any language. We provide qualitative examples of phonetic corruption and recovery in \autoref{fig:pida-qual}.

We focus on Vietnamese as a case study for tonal, isolating languages. Our contributions are: (1) the first phonetic categorization and impact analysis of ASR errors for Vietnamese ST, revealing that most errors are systematic phonetic confusions; (2) phonetically-informed data augmentation using phonetic embeddings motivated by this analysis; (3) demonstration that \textbf{fine-tuning on a mix of clean \& PiDA-corrupted text improves ST (up to +2.04 BLEU) while improving MT performance}. Our method avoids the text translation degradation caused by fine-tuning on real ASR outputs, and requires no audio data. Being text only, PiDA has the potential to be utilized in domains with available NMT data but limited speech resources.

\section{Related Work}

\textbf{Vietnamese Speech Translation.}  
Vietnamese ST remains underrepresented in the literature, with only two data sources providing (audio, transcripts, translation) triplets: the Vietnamese split of multilingual datasets MultiMed-ST \cite{le-duc-etal-2025-multimed-st} (16 hours, 9.1k samples) and FLEURS \cite{conneau2023fleurs} (12 hours, 4.2k samples). This lack of high-quality datasets motivates our use of cascaded ST where ASR \cite{nguyen2026vimedcssvietnamesemedicalcodeswitching, ardila-etal-2020-common, luong-vu-2016-non} and NMT \cite {vo-etal-2024-improving, ngo2022mtetmultidomaintranslationenglish, doan-etal-2021-phomt} modules can be tuned separately with more abundant resources. 

We leverage FLEURS for our ASR error quantification and data augmentation experiments. We also evaluated MultiMed-ST but found substantial audio-transcript misalignments (67 out of 100 manually verified samples contained an average of 2.39 missing and 1.57 erroneously present words per sample) that inflate observed error rates, making the error analysis and speech translation benchmarking on this dataset unreliable.

\textbf{ASR Error Analysis for Machine Translation.}  
\cite{ruiz-federico-2014-assessing} applied linear mixed-effects models to quantify the impact of ASR errors on statistical MT, decomposing errors into substitutions, insertions, and deletions. They found substitutions most detrimental to translation quality, though treated them as one single category. We extend the regression analysis of \cite{ruiz-federico-2014-assessing} by decomposing substitutions according to their phonetic cause.

\textbf{Data Augmentation for Robust Spoken Language Tasks.}  
\cite{di-gangi-etal-2019-robust} fine-tune the NMT component on paired clean and noisy transcripts to improve cascaded ST. \cite{sperber2017toward} show that NMT robustness improves through synthetic noise augmentation when noise type and amount are carefully calibrated. \cite{cheng2018towards} use adversarial training to improve robustness. Recently, MEDSAGE \cite{binici2025medsage} extracts error statistics from real ASR outputs and uses a large language model (LLM) to generate ASR-like corrupted text matching these profiles. While effective for English dialogue summarization, MEDSAGE relies on an LLM's implicit notion of phonetic plausibility, which may inadequately model true acoustic confusability for phonologically distant languages. We instead use phonetic embeddings to identify phonetically similar syllables.

\textbf{Phonetic Word Embeddings.}  
Phonetic representations range from rule-based articulatory features (PanPhon \cite{mortensen-etal-2016-panphon}) to learned phoneme embeddings such as Phoneme2Vec \cite{fang2020phoneme2vec}, with benchmarks like PWESuite \cite{zouhar-etal-2024-pwesuite} and PSET \cite{sperduti-nguyen-2025-pset} evaluating their ability to capture phonetic similarity. More recently, XPhoneBERT \cite{thenguyen23_interspeech}, a multilingual phoneme-level Transformer pretrained on 330M phoneme sequences across nearly 100 languages (including Vietnamese), introduced contextualized phoneme representations. Originally, Phoneme2Vec incorporated phoneme embeddings as an additional input channel in classifier models for ASR-robust sentence classification, whereas XPhoneBERT replaced the phoneme encoder in VITS \cite{pmlr-v139-kim21f} to provide contextualized phoneme representations for Text-to-Speech. We instead adopt the pretrained XPhoneBERT embeddings to model phonetic similarity for data augmentation.

\section{ASR Error Analysis}

\subsection{Error Extraction and Alignment}

We use PhoWhisper-large (\texttt{PhoWhisper-large}) \cite{le2024phowhisper} and wav2vec2-base (\texttt{wav2vec2-base-vietnamese-250h}) \cite{Thai_Binh_Nguyen_wav2vec2_vi_2021}, Vietnamese ASR models based on Whisper \cite{radford2023whisper} and Wav2vec 2.0 \cite{baevski2020wav2vec20frameworkselfsupervised}, to transcribe the Vietnamese training split of FLEURS Vietnamese-English (3k samples).
ASR outputs are word-aligned with reference transcripts using minimum edit distance, from which we extract substitution, insertion, and deletion errors. 

\subsection{Phonetic Categorization of Substitution Errors}

For each substitution error, we classify it into one of six categories based on the phonetic relationship between the reference word and the ASR hypothesis:

\begin{itemize}[nosep,leftmargin=*]
    \item \textbf{Vowel confusion:} The substitution involves changes in vowel quality, vowel diacritics, or vowel length while the consonant structure remains similar. Example: \vn{sau}$_\text{after}$ $\rightarrow$ \vn{sao}$_\text{star}$
    \item \textbf{Consonant confusion:} The substitution involves consonant changes while vowels remain similar. Example: \vn{sử}$_\text{history}$~$\rightarrow$~\vn{xử}$_\text{execute}$. 
    \item \textbf{Tonal confusion:} The reference and hypothesis words are segmentally identical but differ in tone marking. Example: \vn{ma}$_\text{ghost}$ $\rightarrow$ \vn{mà}$_\text{but}$.
    \item \textbf{OOV (Out-of-vocabulary):} The reference contains a foreign term, typically English vocabulary, that the ASR system transcribes as Vietnamese syllables approximating the foreign phonetics. Example: \textit{feet} $\rightarrow$ \vn{phít}.
    \item \textbf{NA (Not applicable):} Substitutions where the reference and hypothesis are phonetically unrelated, possibly due to alignment errors or severe ASR failures.
    \item \textbf{No error:} Orthographic variants representing the same word, mostly due to typography in the reference transcription (e.g., \vn{hoà}$_\text{draw}$ $\rightarrow$  \vn{hòa} or \vn{tỉ}$_\text{billion}$ $\rightarrow$  \vn{tỷ}). These are not semantic errors.
\end{itemize}

Classification is performed using Gemini 2.5 Flash \cite{comanici2025gemini25pushingfrontier} with a prompt containing definitions, decision criteria, and examples for each category. As validation, we manually labeled the top 100 substitution errors and compared against LLM labels, achieving agreement in 94 cases and yielding a Cohen's $\kappa$ = $0.95$. Disagreements primarily involved ambiguous cases where multiple phonetic differences co-occurred. In these cases, the LLM classifier typically assigns the dominant phonetic cause.

\subsection{Impact Quantification via Mixed-Effects Modelling}

Following \cite{ruiz-federico-2014-assessing}, we use a Linear Mixed-Effects Model (LMM) to quantify how each error type impacts translation quality while controlling for differences between ASR systems. We use VinAI-Translate (\texttt{vinai-translate-vi2en-v2}) \cite{nguyen22e_interspeech} (a fine-tuned mBART \cite{liu-etal-2020-multilingual-denoising}, leading model in recent Vietnamese NMT benchmarks \cite{le-duc-etal-2025-multimed-st, vo-etal-2024-improving}) to translate the ground-truth text of FLEURS and the ASR outputs of wav2vec2-base and PhoWhisper-large.

We define the dependent variable in our LMM as:
\begin{equation}
\Delta\text{TER} = \text{TER}_{\text{ST}} - \text{TER}_{\text{MT}}
\end{equation}
where $\text{TER}_{\text{ST}}$ is Translation Edit Rate using ASR output as NMT input and $\text{TER}_{\text{MT}}$ uses reference transcripts. This difference isolates translation degradation specifically attributable to ASR errors. We use TER rather than BLEU following \cite{ruiz-federico-2014-assessing} for its favorable statistical properties in linear modeling.

The mixed-effects model takes the form:
\begin{equation}
    \Delta\text{TER} = \beta_0 + \sum_i \beta_i \cdot x_i + \gamma \cdot \text{ASR}_{\text{sys}} + u_j + \epsilon
\end{equation}
where $x_i$ represents the normalized frequency (count divided by sentence length) of each error type, $\text{ASR}_{\text{sys}}$ is a fixed effect controlling for differences between the two ASR systems, and $u_j \sim \mathcal{N}(0, \sigma^2_u)$ is a random intercept per sentence $j$, accounting for variation in translation difficulty across utterances.

\subsection{Results}

\begin{table}[t]
    \centering
    \caption{Distribution of ASR substitution errors by phonetic cause. PhoWhisper-large and wav2vec2-base achieve 8.62\% and 15.26\% WER on the FLEURS train set, respectively.}
    \label{tab:error_dist}
    \begin{tabular}{lrr}
        \toprule
        \textbf{Error Type} & \textbf{PhoWhisper} & \textbf{wav2vec2} \\
        \midrule
        OOV (foreign terms) & 2,612 & 3,053 \\
        Vowel confusion & 566 & 1,627 \\
        Consonant confusion & 447 & 1,423 \\
        NA (phonetically unrelated) & 533 & 803 \\
        Tonal confusion & 175 & 388 \\
        No error (orthographic variant) & 365 & 104 \\
        \midrule
        \textbf{Total substitutions} & 4,698 & 7,398 \\
        \bottomrule
    \end{tabular}
\end{table}

\begin{table}[t]
    \centering
    \caption{Mixed-effects regression coefficients predicting $\Delta$TER. Error counts are length-normalized. 
    *: $p<0.05$}
    \label{tab:regression}
    \begin{tabular}{lrrc}
        \toprule
        \textbf{Predictor} & \textbf{Coef.} & \textbf{Std. Err.} & \textbf{Sig.} \\
        \midrule
        Intercept & 0.74 & 0.35 & * \\
        Deletions & 133.95 & 11.45 & * \\
        Insertions & 127.51 & 2.47 & * \\
        Vowel conf. & 97.66 & 7.47 & * \\
        Tonal conf. & 79.12 & 14.27 & * \\
        Consonant conf. & 71.84 & 8.07 & * \\
        OOV & 58.82 & 5.72 & * \\
        NA & 23.19 & 14.01 &  \\
        \bottomrule
    \end{tabular}
\end{table}

\autoref{tab:error_dist} shows the error distribution across both ASR systems. Among within-vocabulary errors, \textbf{phonetic confusions are most prevalent}: vowel and consonant confusions together account for 3,050 errors in wav2vec2 and 1,013 in PhoWhisper, outnumbering tonal confusions and phonetically unrelated substitutions. 

\autoref{tab:regression} presents mixed-effects coefficients quantifying translation impact\footnote{\textit{No error} is excluded because its coefficient is statistically insignificant in a preliminary model, and it does not represent semantic errors.}. Insertions and deletions show the largest effects, consistent with prior work~\cite{ruiz-federico-2014-assessing}. Among substitution types, \textbf{phonetic confusions have the most dominant impacts} on ST degradation. Vowel confusions have the strongest effect (97.66), followed by tonal (79.12) and consonant confusions (71.84), while phonetically unrelated errors (NA) are insignificant.

\section{Phonetically-Informed Data Augmentation}
We propose the \textbf{P}honetically-\textbf{I}nformed \textbf{D}ata \textbf{A}ugmentation (\textbf{PiDA}) pipeline, consisting of six components across two phases:

\subsection{Precomputation Phase}
\textbf{1. Syllable Inventory Construction.} We extract the Vietnamese syllable inventory using the \texttt{wordfreq} library \cite{robyn_speer_2022_7199437}, which provides frequency statistics derived from large web corpora\footnote{Vietnamese has a finite, constrained syllable inventory, and web corpora like \texttt{wordfreq} cover nearly all attested syllables.}. We retrieve the top 50,000 most frequent Vietnamese words and filter to entries matching valid Vietnamese orthographic patterns, yielding approximately 9,400 unique syllables.

\textbf{2. Phoneme Conversion.} Each syllable is converted to International Phonetic Alphabet (IPA) using CharsiuG2P \cite{zhu22_interspeech}, the grapheme-to-phoneme system used by XPhoneBERT.

\textbf{3. Embedding Extraction.} We pass each phoneme sequence through XPhoneBERT (\texttt{xphonebert-base}) \cite{thenguyen23_interspeech} and mean-pool the output hidden states across all phoneme positions (excluding the \texttt{[CLS]} and \texttt{[SEP]} tokens) to obtain a 768-dimensional vector per syllable. 

\textbf{4. Similarity Index Construction.} We L2-normalize all syllable embeddings and build a FAISS index \cite{johnson2019billion} using inner product search for efficient approximate nearest-neighbor retrieval. For each syllable, we precompute its top-50 phonetically similar neighbors and their cosine similarity scores.

\subsection{Augmentation Phase}

\textbf{5. Error Annotation.} We follow the error annotation procedure of \cite{binici2025medsage}: Given an input sentence to corrupt, we annotate individual words with error markers based on ASR error statistics computed from the training set. For each word $w_i$, we sample from a Bernoulli distribution with probability $p = \text{WER}_{\text{train}}$ (the observed word error rate on the training set). If selected for corruption, we sample an operation type (deletion or substitution) according to the observed proportions of these error types in real ASR errors\footnote{We exclude insertion simulation as our pipeline lacks a language model component to generate contextually plausible insertions.}. We mark substitutions with braces $\{w_i\}$ and deletions by omitting the word entirely.

\textbf{6. Corruption.} We then process the annotated text to generate the corrupted output. For deletion markers, the word is simply removed. For substitution markers, we perform phonetic corruption with the following procedure. 

For each syllable $s_i$ marked for substitution, we sample a replacement from its top-$k$ ($k=5$) precomputed phonetic neighbors using temperature-scaled softmax sampling:

\begin{equation}
P(s_j \mid s_i) = \frac{\exp\bigl(\text{sim}(s_i, s_j) / \tau\bigr)}{\sum_{s_k \in \mathcal{N}_k(s_i)} \exp\bigl(\text{sim}(s_i, s_k) / \tau\bigr)}
\end{equation}

where $\text{sim}(s_i, s_j)$ is the cosine similarity between syllable embeddings, $\mathcal{N}_k(s_i)$ denotes the top-$k$ neighbors of $s_i$, and $\tau$ is a temperature parameter controlling the sharpness of the distribution (lower values favor more phonetically similar substitutions).

\begin{table*}[t]
    \centering
    \caption{Translation quality on the FLEURS Vi-En test set for clean text translation (MT) and speech translation from ASR transcripts (ST).
    Each setting reports BLEU and COMET.
    $^{+}$ / $^{-}$: statistically significant BLEU increase/decrease over baseline (paired bootstrap, $p < 0.05$).
    Best and second-best BLEU in \textbf{bold} and \underline{underline}.
    $\checkmark$ indicates text-only / LLM-independent method.}
    \label{tab:results}
    \begin{tabular}{l cc cc cc cc}
        \toprule
        \textbf{Data Aug. Method} 
        & \textbf{Text-only}
        & \textbf{LLM-indep.}
        & \multicolumn{2}{c}{\textbf{MT}} 
        & \multicolumn{2}{c}{\textbf{ST (PW)}} 
        & \multicolumn{2}{c}{\textbf{ST (w2v)}} \\
        \cmidrule(lr){4-5} \cmidrule(lr){6-7} \cmidrule(lr){8-9}
        & & 
        & BLEU$\uparrow$ & COMET$\uparrow$
        & BLEU$\uparrow$ & COMET$\uparrow$
        & BLEU$\uparrow$ & COMET$\uparrow$ \\
        \midrule
        No fine-tuning 
        & 
        & 
        & 28.05 & 0.84 
        & 23.73 & 0.80 
        & 22.15 & 0.73 \\
        + clean pairs (\textit{baseline}) 
        & $\checkmark$
        & $\checkmark$
        & 33.04 & 0.86 
        & 26.25 & 0.81
        & 22.40 & 0.75 \\
        \midrule
        + clean \& freq-based subs
        & $\checkmark$
        & $\checkmark$
        & 33.13 & 0.86
        & 27.45$^{+}$ & 0.81
        & 22.77 & 0.77 \\
        + freq-based subs only
        & $\checkmark$
        & $\checkmark$
        & 32.90 & 0.86
        & 27.46$^{+}$ & 0.82
        & 22.52 & 0.78 \\
        \midrule
        + clean \& real noisy 
        & 
        & $\checkmark$
        & 33.09 & 0.86 
        & 27.43$^{+}$ & 0.82 
        & 22.96 & 0.76 \\
        + real noisy only 
        & 
        & $\checkmark$
        & 32.00$^{-}$ & 0.85 
        & 28.06$^{+}$ & 0.83 
        & \textbf{23.65}$^{+}$ & 0.78 \\
        \midrule
        + clean \& MEDSAGE 
        & $\checkmark$
        & 
        & 32.59 & 0.86 
        & 26.68 & 0.82 
        & 22.88 & 0.77 \\
        + MEDSAGE only 
        & $\checkmark$
        & 
        & 32.45$^{-}$ & 0.86 
        & 26.43 & 0.82 
        & 22.48 & 0.78 \\
        \midrule
        + \textit{clean \& PiDA (ours)} 
        & $\checkmark$
        & $\checkmark$
        & \textbf{33.72}$^{+}$ & 0.86 
        & \textbf{28.29}$^{+}$ & 0.82 
        & 23.18$^{+}$ & 0.77 \\
        + \textit{PiDA only (ours)} 
        & $\checkmark$
        & $\checkmark$
        & \underline{33.58} & 0.86 
        & \underline{28.19}$^{+}$ & 0.82 
        & \underline{23.43}$^{+}$ & 0.77 \\
        \bottomrule
    \end{tabular}
\end{table*}

\section{Experiments}

\subsection{Experimental Setup}

\textbf{Data.} We use PiDA to corrupt the training split of FLEURS, then evaluate the fine-tuned models on 0.9k test samples. 

\textbf{Models.} We use PhoWhisper-large and wav2vec2-base  as the ASR systems, and VinAI-Translate as the NMT. 

\textbf{Training protocol.} We fine-tune VinAI-Translate with: 3 epochs, batch size 8, learning rate $3\times10^{-5}$, maximum sequence length 256 tokens, gradient accumulation over 8 steps (effective batch size 64), 300 warmup steps, weight decay 0.01, AdamW optimizer with early stopping based on validation loss.


\textbf{Baselines.} We compare against: (1) \textit{Clean pairs}: baseline fine-tuning on reference Vietnamese-English pairs only; (2) \textit{Random frequency-based substitutions} \cite{sperber2017toward}: corrupting reference transcripts by sampling error operations (deletion, substitution) according to WER statistics from real ASR outputs, with substitutions drawn from frequency-weighted Vietnamese syllable inventory (the same as PiDA's); (3) \textit{Real noisy pairs} \cite{di-gangi-etal-2019-robust}: using actual ASR outputs paired with English translations; (4) \textit{MEDSAGE} \cite{binici2025medsage}: similarly sampling error operations from ASR error profiles, but using LLM-generated insertions and substitutions via Gemini 2.5 Flash instead of random vocabulary draws\footnote{MEDSAGE originally used Llama-3-8B \cite{grattafiori2024llama3herdmodels} and Mistral-7B \cite{jiang2023mistral7b} for corruption. However, we found that their Vietnamese fine-tuned counterparts, Llama-SEA-LION-v3.5-8B-R \cite{sealion_v3_5_8b_r} and Vistral-7B \cite{chien2023vistral}, failed to produce plausible phonetic errors for Vietnamese, even with few-shot examples. Hence, we used Gemini 2.5 Flash for this component.}. In all conditions, when clean and noisy data are mixed (denoted as \textit{clean \& \{method name\}}), the ratio is always 1:1 (3k clean \& 3k noisy), following \cite{di-gangi-etal-2019-robust}.

\textbf{Evaluation.} We report BLEU \cite{papineni-etal-2002-bleu} and COMET \cite{rei-etal-2022-comet} (\texttt{Unbabel/wmt22-comet-da}) on clean reference transcripts (MT) and on PhoWhisper-large and wav2vec2-base ASR outputs (ST). We abbreviate the metrics as \texttt{Metric}$_\texttt{Setting}^\texttt{Model}$. For example, BLEU$_\text{ST}^\text{PW}$ refers to the BLEU score for speech translation using the PhoWhisper-large ASR outputs.

\subsection{Results} \label{experimental_results}


\autoref{tab:results} presents our main results across training conditions.

\textbf{Baseline and ASR error propagation.} Fine-tuning on clean reference FLEURS pairs improves speech translation quality, but a substantial MT-ST gap remains (6.79 BLEU for PhoWhisper-large and 10.64 BLEU for wav2vec2-base), indicating that domain adaptation alone does not address robustness to ASR errors, particularly for weaker ASR front-end.


\textbf{Random substitutions \& Real noisy pairs.} Frequency-based random corruption yields ST gains (27.45--27.46 BLEU$_\text{ST}^\text{PW}$, $p < 0.05$) comparable to mixing clean with real noisy text (27.43 BLEU$_\text{ST}^\text{PW}$, $p < 0.05$) without degrading BLEU$_{\text{MT}}$.  Fine-tuning with only real noisy pairs increases both BLEU$_\text{ST}^\text{PW}$ (28.06) and BLEU$_\text{ST}^\text{w2v}$ (23.65), both $p < 0.05$. However, noisy-only training \textbf{significantly degrades BLEU$_{\text{MT}}$} by 1.04 ($p < 0.05$).

\textbf{MEDSAGE.} LLM-generated corruptions yield no significant BLEU improvement over the clean baseline for either ASR system, though COMET scores are similar to other methods. 

\textbf{PiDA (ours).} Mixing clean text with PiDA achieves the strongest BLEU$_\text{ST}^\text{PW}$ (+2.04 over baseline, +0.84 over random substitutions, $p < 0.05$) and significant gains in BLEU$_\text{ST}^\text{w2v}$ (+0.78, $p < 0.05$), while preserving both BLEU\textsubscript{MT} and COMET\textsubscript{MT}. Real-noisy-only training achieves slightly lower BLEU$_\text{ST}^\text{PW}$ and slightly better BLEU$_\text{ST}^\text{w2v}$, but degrades BLEU\textsubscript{MT}. \textbf{Mixing clean data \& PiDA is the only method that improves speech translation across both ASR systems while also improving clean-text performance.} Without clean pairs, PiDA still preserves MT quality and delivers significant ST gains. COMET scores across all augmentation methods are comparable.

Overall, PiDA consistently provides robustness gains comparable to real noisy data while eliminating the robustness-accuracy tradeoff, and also requiring neither audio data nor external LLMs for text augmentation. 

 \textbf{Sampling hyperparameter ablation.}
We analyze the sensitivity of PiDA to the sampling hyperparameters $k$ and $\tau$ in \autoref{tab:k_tau_ablation}, showing that performance is stable across a wide range of values. The best BLEU$_\text{ST}^\text{{PW}}$ and BLEU$_\text{ST}^\text{{w2v}}$ are achieved with $k{=}5$ and $\tau{=}0.5$, which we used in our main experiments.

\begin{table}[t]
    \centering
    \caption{Ablation for hyperparameters $k \in \{3,5,10\}$ and $\tau \in \{0.3, 0.5, 1.0\}$ on FLEURS Vi-En. Each cell reports \textup{BLEU$_\text{ST}^\text{PW}$ | BLEU$_\text{ST}^\text{w2v}$} of models fine-tuned on `clean \& PiDA' data. 
    }
    \label{tab:k_tau_ablation}
    \begin{tabular}{c|ccc}
        \toprule
        \textbf{$k \,|\, \tau$} & ${0.3}$ & ${0.5}$ & ${1.0}$ \\
        \midrule
        \textit{$3$}  
        & 27.62 | 22.93         & 27.74 | 23.00         & 27.76 | 22.91 \\
        \textit{$5$}  
        & 27.53 | 23.11         & \textbf{28.29} | \textbf{23.18}         & 27.35 | 23.11 \\
        \textit{$10$} 
        & 27.92 | 22.91         & 27.96 | 22.91         & 27.83 | 22.93 \\
        \bottomrule
    \end{tabular}
\end{table}

\section{Conclusion \& Future Work}

We presented the first systematic categorization of ASR errors for Vietnamese ST, showing that most substitution errors stem from structured phonetic confusions, and that these errors substantially affect downstream NMT performance. 

This motivated \textbf{P}honetically-\textbf{I}nformed \textbf{D}ata \textbf{A}ugmentation using XPhoneBERT embeddings (\textbf{PiDA}), which generates synthetic errors by substituting phonetically similar syllables. Across ASR systems, PiDA achieves consistent gains in ST, scoring the highest BLEU$_\text{ST}$ when translating PhoWhisper-large outputs and significant improvements for  wav2vec2-base, while retaining or improving text translation quality on both BLEU and COMET. In contrast, real-ASR augmentation yields comparable ST gains but degrades MT performance. Being text-only, PiDA has the potential to extend to ST domains where NMT parallel data exists but domain-specific speech resources are limited.

\textbf{Limitations} We only evaluate on one dataset due to the lack of high-quality ST datasets for Vietnamese. OOV errors involve cross-lingual phonetic mapping not addressed by our within-vocabulary method. Future work will extend to OOV handling and validate on additional Vietnamese ST benchmarks.

\section{Acknowledgments}
The research results are a part of the outputs of the Cross-College project \textbf{\textit{Robust Vietnamese–English Clinical and Educational Medical Translation}}, a collaboration between the College of Engineering \& Computer Science (CECS) and the College of Health Sciences (CHS), VinUniversity. This research was funded under Project ID VUNI.2324.CC06. 

Giang Son Nguyen, Tung X. Nguyen, Hieu Minh Truong and Dung D. Le are partly supported by the Center for AI Research, VinUniversity. Nhu Vo is supported by the Vingroup Scholarship.

Giang Son Nguyen thanks the AUMOVIO-NTU Corporate Lab for lending GPU resources for the experiments in this paper.

\section{Generative AI Use Disclosure}
The authors used generative AI tools solely to assist with minor language editing and readability improvements. No generative AI tools were used to produce any scientific content, experimental results, data analysis, or conclusions.

\bibliographystyle{IEEEtran}
\bibliography{references}

\end{document}